\documentclass[letterpaper, 10 pt, journal, twoside]{ieeetran}
\usepackage{marginnote}
\usepackage{graphics} % for pdf, bitmapped graphics files
\usepackage{graphicx}   % To include figures

\usepackage{amsmath} % assumes amsmath package installed
\usepackage{mathtools, cuted}
\usepackage{interval}
\usepackage[noadjust]{cite}
\usepackage{siunitx}
\usepackage{breqn}
\usepackage{lscape}
\usepackage{derivative}
\usepackage{nicematrix}
\usepackage{color, colortbl}
\usepackage{multirow}
\usepackage{nicefrac}
\usepackage{amssymb}  % assumes amsmath package installed
\usepackage{hyperref}

\usepackage{tabularx}

\newtheorem{definition}{Definition}

\renewcommand{\vec}[1]{\mathbf{\boldsymbol{#1}}}

\newcommand{\de}{{\mathrel{\mathop:}=}}

\newcommand{\etal}{\textit{et al}. }
\newcommand{\ie}{\textit{i}.\textit{e}., }

\providecommand{\R}{\ensuremath \mathbb{R}}

\providecommand{\Z}{\ensuremath \mathbb{Z}}

\providecommand{\Review}[3]{#2}
\providecommand{\ReviewL}[3]{#2}

\begin{document}

\author{Jiayu Ding, Zhenyu Gan
\thanks{Manuscript received: November 27, 2023; Revised February 8, 2024; Accepted March 16, 2024. This paper was recommended for publication by Editor Abderrahmane Kheddar upon evaluation of the Associate Editor and
Reviewers’ comments.}
\thanks{Jiayu Ding and Zhenyu Gan are with the Mechanical and Aerospace Engineering, Syracuse University, Syracuse, NY 13244 \texttt{\{jding14, zgan02\}@syr.edu }. This work was supported by a startup fund from the Syracuse University.}
\thanks{Digital Object Identifier (DOI): 10.1109/LRA.2024.3384908.}
}

% Paper headers
% \markboth{IEEE ROBOTICS AND AUTOMATION LETTERS. PREPRINT VERSION. ACCEPTED March, 2024}
% {Ding \MakeLowercase{\textit{et al.}}: Breaking Symmetries Leads to Diverse Gaits} 

\markboth{ACCEPTED by IEEE RA-L, March, 2024. For citation, please use DOI: 10.1109/LRA.2024.3384908.}
{Ding \MakeLowercase{\textit{et al.}}: Breaking Symmetries Leads to Diverse Gaits} 

\title{Breaking Symmetries Leads to \\ Diverse Quadrupedal Gaits}
% \begin{document}

\maketitle

\begin{abstract}
Symmetry manifests itself in legged locomotion in a variety of ways:
A legged system can maintain consistent gaits from any spatial starting point, exhibiting the same leg movements on either side of the torso in phase, and some even demonstrate forward and backward movements so similar they seem to reverse time.
This work aims to generalize these phenomena and proposes formal definitions of symmetries in legged locomotion using terminology from group theory.
In this research, we uncovered an intrinsic connection among a broad spectrum of quadrupedal gaits, which can be systematically identified via numerical continuations and distinguished by elements within a symmetry group. These gaits, within the hybrid dynamical system, are not merely isolated movements but part of a continuum, seamlessly transitioning from one to another at precise parameter bifurcation points. Altering specific symmetries at these junctures leads to the emergence of distinct gaits with unique footfall patterns, a phenomenon we've generalized through dimensional analysis in this study. 
Consequently, each gait manifests distinct preferred speed ranges and specific transition speeds.
This work offers a comprehensive method to solve the gait generation problem for a quadruped, including pronking, two types of bounding, four variations of half-bounding, and two forms of galloping, and it also elucidates the mechanical rationale behind the necessity of gait transitions, 
providing high-level insight into the diversity and underlying mechanics of quadrupedal locomotion.
\end{abstract}


\section{Introduction}
\IEEEPARstart{L}{egged} locomotion has become an increasingly prominent area of research,  as it has numerous applications in the areas of biomechanics and robotics.
Scientists and engineers can develop better mobile robots by studying the dynamics and control of animals as they transverse difficult terrains in nature, for example.
Among all these studies, a considerable amount of attention has been paid to the periodic motion patterns of legged systems, also known as \emph{gaits}.
Each gait is observed to have a unique set of properties, such as various leg movement patterns and oscillation modes of the body, which can vary rapidly according to terrain, speed, and body mass.
For example, during the \emph{trotting} gait of a quadruped, the motion of the diagonal pairs of legs is fully synchronized, and the torso barely rotates during the entire stride.
Because of its simplicity, trotting gait is widely adopted on quadrupedal robots such as the MIT mini-cheetah robot \cite{katz2019mini} or the ETH ANYmal robot \cite{hutter2016anymal}.
In contrast, animals may use \emph{bounding} gaits, which synchronize the motion within the front and hind leg pairs, to jump across obstacles when limited space can be deployed between footholds; or use \emph{galloping}, which are characterized by a rapid succession of footfalls followed by a prolonged flight phase, to attain their maximum speeds.
It has been hypothesized that each gait has its own optimal speed, and they play a key role in maintaining balance and saving energy \cite{Hoyt1981gaitenergetic}.
Similarly to changing the gears of the transmission system in a car, by altering their gaits, legged systems have the potential to travel longer distances or cover difficult terrains in a shorter time.
However, due to the high degrees of freedom in legged systems and the intermittent ground contact with large impulsive forces, studying legged locomotion and gait patterns remains challenging. 
In spite of the fact that it is possible to construct a hybrid model that can reproduce many common quadrupedal gaits \cite{gan2016passive}, it is not computationally feasible to identify every possible gait due to the high dimensionality of the model.

\begin{figure}[tpb]
\centering
\includegraphics[width=1\columnwidth]{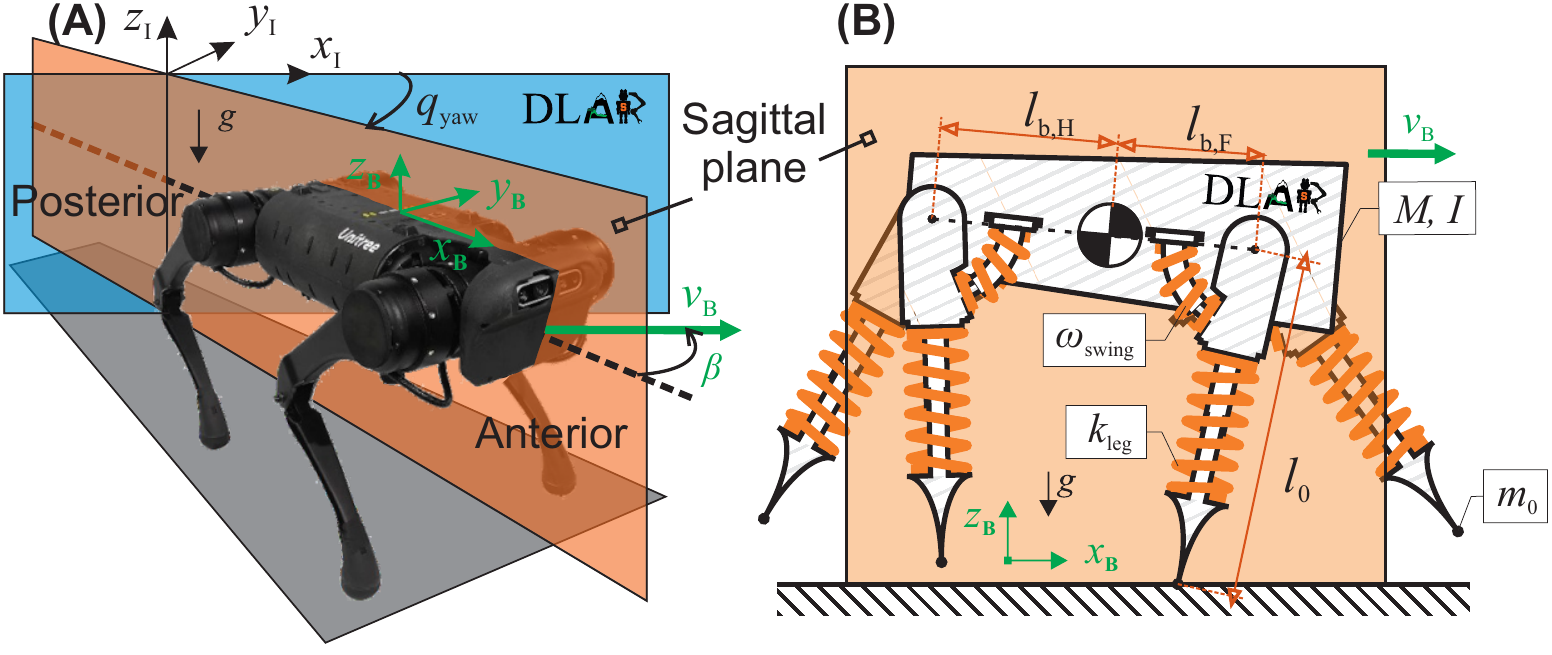}
\caption{(A) A1 quadrupedal robot from Unitree Robotics with bilateral symmetry and (B) a simplistic spring-mass model.}
\label{fig:model}
\vspace*{-0.2in}
\end{figure}

The concept of \emph{symmetries} has been widely used in the research of legged locomotion in order to reduce the complexity of the study, reveal fundamental differences among all gaits, and design locomotion controllers for legged robots. 
For example, Hildebrand (1965) used the phase delays among the leg pairs (front or hind) to categorize all the quadrupedal gaits \cite{Hildebrand1965}.
If the phase delay in the gait is equal to half of the stride cycle, he called it a \emph{symmetrical gait}; otherwise, an \emph{asymmetrical gait}. 
He suggested that all the symmetrical gaits and asymmetrical gaits formed two distinct continua \cite{Hilderbrand1989Quadrupedal} and animals like horses can smoothly switch from one gait to another if they are closely related.
Another type of symmetry was brought up by Marc Raibert (1986) from his early studies of running legged robots, in which he found when the robot was moving backward, it was almost the same as moving forward in a time-reversed fashion \cite{Raibert1986symmetry}.
More recently, Razavi \etal (2017) introduced a new method to design stable periodic walking gaits for legged robots based on producing a type of odd-even symmetry in the system \cite{razavi2017symmetry}. 
This approach can identify symmetries in the system dynamics and generate periodic walking without relying on any kind of offline or online numerical search.
Additionally, the study of symmetries has been proven useful for data-driven analysis of robotic systems \cite{ordonez2023discrete}.
By recognizing and exploiting data symmetries through the use of invariant neural networks (NN), symmetry constraints can lead to better sample efficiency and reduce the number of trainable parameters.

In this study, we try to leverage the existing studies of symmetries in legged locomotion \cite{Razavi2016} and group theory \cite{razavi2017symmetry} and seek to understand the inherent relationships among the large number of asymmetrical gaits observed by Hildebrand \cite{Hildebrand1977}.
The contributions of our study are as follows. 
Firstly, we provide formal definitions for each symmetry type found in the existing literature, utilizing a generalized floating base model for the legged system, thereby uncovering the intrinsic relationships among these symmetries. 
Secondly, by using a simplified model, we illustrate how breaking symmetries gave rise to novel gait patterns and how various quadrupedal gaits can be systematically identified and interconnected. These findings \Review{suggest}{confirm}{9-4} that gaits are oscillation modes of a hybrid dynamical system and offer a \Review{foundational }{}{9-4} framework for a generalized gait generation methodology that is versatile enough to be adapted across a wide range of quadruped robots with different mechanical designs.
Lastly, we generalize all the results using a dimensionless analysis and conduct parameter studies, including variations in the torso's center of mass (COM) location, to illuminate the diverse mechanisms by which symmetries are disrupted.

\section{Methods}
Symmetries of mechanical systems are natural examples of group actions, and they often provide a useful mechanism to reduce the complexity of the problem by reducing the number of variables.
In this section, we seek to generalize the commonly discussed symmetries in a legged system and provide a systematic approach to analyze the symmetries of a specific gait pattern.
Furthermore, we introduce a simplistic model that can reproduce a large number of quadrupedal gaits and use it to demonstrate how symmetries break through parameter bifurcations.

\subsection{Models of Legged systems}
A quadrupedal legged system can be approximated by a floating-base model (FBM) consisting of rigid bodies with mass and inertia, connected via joints to form an open kinematic chain in 3-dimensional Euclidean space. 
Let $\mathcal{Q}$ be the $n$-dimensional configuration space of the robot.
We use variables $\vec{q}_{T} \de \left[ q_x,q_y,q_z,q_{\text{yaw}},q_{\text{pitch}},q_{\text{roll}} \right]^\intercal$ to represent the Cartesian position of the torso's geometrical center in the inertial coordinate frame (I) and the torso’s intrinsic Euler angles in the $z{\text -}y{\text -}x$ order, respectively.
For the $i$-th leg, the joint vector (shape coordinates) $\vec{q}_{i}$ refers to the relative angles of the joints measured in their own body coordinate frames.
The index $ i \in L = {\{\footnotesize\text{LH}, \text{LF}, \text{RF}, \text{RH}\}}$ stands for the left hind, the left front, the right front, and the right hind legs.
By collecting the leg configurations $\vec{q}_{L} \de \left[ \vec{q}_{\text{LH}}^\intercal, \vec{q}_{\text{LF}}^\intercal, \vec{q}_{\text{RF}}^\intercal, \vec{q}_{\text{RH}}^\intercal \right]^\intercal$,
the generalized coordinates of the FBM are aggregated in a single configuration space vector  $\vec{q} \de \left[\vec{q}_{T}^\intercal, \vec{q}_{L}^\intercal \right] ^\intercal$. 
By applying the Euler-Lagrange equation, the equations of motion of a legged robot can be expressed in the following form:
\begin{equation}
\label{eq:EOM}
   \operatorname{\vec{M}}(\vec{q})\ddot{\vec{q}} + \operatorname{\vec{C}}(\vec{q},\dot{\vec{q}})\dot{\vec{q}} + \operatorname{\vec{G}}(\vec{q}) = \vec{S}\vec{\tau}  + \vec{J}^\intercal(\vec{q})\vec{\lambda} .
\end{equation}
where $\operatorname{\vec{M}}(\vec{q})$ is the inertia matrix; $\operatorname{\vec{C}}(\vec{q},\dot{\vec{q}})$ is the Coriolis matrix;
and $\operatorname{\vec{G}}(\vec{q})$ is the gravitational vector. 
$\vec{\tau}$ denotes the vector of joint motor torques and $\vec{S}$ is the selection matrix that assigns motor torques to the generalized coordinates.
When a leg is in stance, the resulting ground reaction force $\vec{\lambda}$ is mapped to the joints through Jacobi mapping $\vec{J}^\intercal(\vec{q})$.

In order to reveal symmetries in a dynamical system, we developed a quadrupedal spring-loaded inverted pendulum (SLIP) model, which consists of a rigid torso with mass $M$ and pitching inertia $I$ and four legs with a resting length $l_o$ and concentrated mass $m_{o}$ at foot position as shown in Fig.\ref{fig:model}(B). 
The COM of the torso lies on the line between the shoulder and hip joints. 
It is located at a normalized distance of $l_{b,H}$ measured from the hip joint and a distance of $l_{b,F}$ from the shoulder joint. 
The length of the torso is set to be equal to the resting leg length $l_o$.
The legs are modeled as linear springs with a stiffness $k_{\text{leg}}$ along the leg.
They are connected to the torso through additional torsional springs with oscillation frequency $\omega_{\text{swing}}$.
In this way, the legs can naturally swing back and forth during flight and there is no need to control the angle of attack for landing with additional parameters.
More information about this process can be found in our previous work on modeling bipedal locomotion \cite{Gan2018allcommonbipedal}, which has been omitted here.
Without loss of generality, we focus on the motion in the sagittal plane by reducing the configuration space to $\vec{q}_T = [q_x,q_z,q_{\text{pitch}}]$ and $\vec{q}_L = [\alpha_{\text{LH}},\alpha_{\text{LF}},\alpha_{\text{RF}},\alpha_{\text{RH}}]$ where $\alpha_i$ are leg angles measured with respect to the torso.

While in flight phases, the legs are maintained at the resting length $l_o$ and the torso is subjected to free fall.
When a leg is in contact with the ground, we assume the horizontal position of a foot $p^x_i$ is a constant value (no sliding motions).
The values of leg angles $\alpha_i$ and leg lengths $l_{i}$ can be computed implicitly from the following kinematic constraints:
\begin{align}
\label{eq:Event_Handler}
 l_{i}(\vec{q})  &=  \left[q_z - l_{b, j} \sin(q_{\text{pitch}})\right] \cos^{-1}(\alpha_i + q_{\text{pitch}}), \\  \quad \notag
 p^x_i(\vec{q})  &=  q_x  - l_{b, j}\cos(q_{\text{pitch}}) + l_{i}(\vec{q}) \sin(\alpha_i + q_{\text{pitch}}) .
\end{align}
where $i \in {\{\footnotesize\text{LH}, \text{LF}, \text{RF}, \text{RH}\}}$, indicating the index of legs, and $j \in {\{\footnotesize\text{H}, \text{F}\}}$, indicating the COM distance from the hind or the front joint.
The contact \Review{jacobian}{Jacobian}{11-5} matrix of the $i$-th stance leg can be calculated as {\footnotesize$\vec{J}_i \mathrel{\mathop:}=  \pdv{\vec{g}_i}{\vec{q}}$} where $\vec{g}_i \de \left[l_{i}(\vec{q}), \: p^x_i(\vec{q}) \right]^\intercal$.
By combining \eqref{eq:EOM} and the augmented contact \Review{jacobian}{Jacobian}{11-5} matrix $\vec{J}_{k}^\intercal = \left[ \vec{J}_{\text{FR}} ^\intercal, \vec{J}_{\text{FL}} ^\intercal, \cdots \right]$ for all stance legs, we obtain $k \in \{1, 2, \ldots,  16\}$ possible domains for the quadrupedal model:
\begin{equation}
\label{eq:DAE} 
{\footnotesize\mathcal{F}_{k} \de
%\resizebox{.9 \linewidth}{!}{$  
    \begin{bmatrix}
           \operatorname{\vec{M}}(\vec{q}) & -\vec{J}_{k}^\intercal(\vec{q}) \\
           \vec{J}_{k}(\vec{q}) &  \vec{0}
    \end{bmatrix}
     \begin{bmatrix}
           \ddot{\vec{q}} \\
           \vec{\lambda}_{k}
    \end{bmatrix}
    = \begin{bmatrix}
           - \operatorname{\vec{C}}(\vec{q},\dot{\vec{q}})\dot{\vec{q}} - \operatorname{\vec{G}}(\vec{q})  \\
           -\vec{\dot{J}}_{k}(\vec{q}, \dot{\vec{q}})\dot{\vec{q}} 
    \end{bmatrix} , }
\end{equation}

where $\vec{\lambda}_k = \left[ \vec{\lambda}_{\text{FR}} ^\intercal, \vec{\lambda}_{\text{FL}} ^\intercal,  \cdots \right]^\intercal$ contains the horizontal and vertical ground reaction forces of all stance legs.

It is assumed in this work that each leg must touch down (TD) and lift off (LO) the ground only once within one stride.
This assumption further implies that Zeno behavior cannot occur in a proper motion from our model \cite{pace2017piecewise, ames2005sufficient}.
The timings of these two events for the $i$-th leg are denoted by $t^{\text{TD}}_{i}$ and $t^{\text{LO}}_{i}$, respectively.
These eight timing variables (two for each leg) are functions of the states ($\vec{q},\vec{\dot{q}}$), and the impact surfaces are defined by the following sets: 
\begin{equation}
\begin{split}
    \label{eq:TD}
    {\footnotesize \mathcal{C}^{\text{TD}}_i(t) = \bigl\{(\vec{q},\vec{\dot{q}}) \in \mathcal{TQ}~\big|~ t = t^{\text{TD}}_{i}(\vec{q},\vec{\dot{q}}), \: l_{i}(\vec{q}) = l_o , \dot{l}_{i}(\vec{q})<0 ~\bigl\}} \\ 
    {\footnotesize \mathcal{C}^{\text{LO}}_i(t) = \bigl\{(\vec{q},\vec{\dot{q}}) \in \mathcal{TQ}~\big|~ t = t^{\text{LO}}_{i}(\vec{q},\vec{\dot{q}}), \: l_{i}(\vec{q}) = l_o  , \dot{l}_{i}(\vec{q})>0 ~\bigl\} }
\end{split}
\end{equation}
that naturally divide all motions into 9 domains ($j \in \{1, 2, \ldots,  9\}$):
\begin{equation} 
\label{eq:HM}
\Sigma(t):   \left\{ 
    \begin{array}{rl}
         \mathcal{F}_{j}, &\hspace{-0.2cm}\left( \vec{q},\dot{\vec{q}} \right) \not \in \mathcal{C}_{j}(t); \\
         \dot{\vec{q}}^+ = \Delta_{\mathcal{F}_{j} \rightarrow \mathcal{F}_{j=j+1}} \dot{\vec{q}}^-, &\hspace{-0.2cm}\left( \vec{q},\dot{\vec{q}} \right) \in \mathcal{C}_{j}(t); \\
    \end{array}
 \right . %$}
\end{equation}
in which $\Delta$ is the impact map that instantaneously resets the pre-impact velocity $\dot{\vec{q}}^-$ joint velocities to the post impact velocities $\dot{\vec{q}}^+$ for the next domain \cite[Chapter~3]{westervelt2018feedback}.

\subsection{Quadrupedal Gaits and symmetries}
As described in the above section, the motion of a legged system can be modeled as a hybrid system $\Sigma$ that consists of a set of differential-algebraic equations $\mathcal{F}$ describing the continuous dynamics of the robot in each domain and a set of impact maps $\Delta$ that instantaneously reset the velocities. 
\ReviewL{Due to the legged system's high degrees of freedom and its hybrid nature, there exist infinite many solutions.}{}{11-4} The \Review{general}{global}{11-4} behavior of such systems has been explored in the literature, notably in \cite{pace2017piecewise, razavi2017symmetry}. In this work, in order to focus on quadrupedal locomotion with diverse footfall patterns, we \Review{define a gait of a legged robot as just a periodic motion of such a system as defined below:}{conduct a local symmetry analysis on the periodic solutions of such systems. A gait of a legged system is defined as:}{11-4}
\vspace{1mm}
\begin{definition}[Gait]
\label{def:periodic motion}
Suppose the equations of motion of a quadrupedal robot can be modeled as a continuous time ($t \in \R$) hybrid system $\Sigma(t)$ on phase space $\mathcal{X}$, where $\vec{x} \in \mathcal{X}, \vec{x}(t) \de \left[\vec{q}(t)^\intercal,  \vec{\dot{q}}(t)^\intercal  \right]^\intercal$. 
It is also assumed that in each stride every leg is limited to a single strike and lift-off from the ground.
The dynamics of $\Sigma(t)$ is given by an evolution operator $\vec{\varphi}_t (\mathcal{X} \rightarrow \mathcal{X}): \vec{x}(\tau) \mapsto \vec{x}(\tau + t)$.
We consider a \emph{gait} of a legged robot as a periodic orbit  $\mathcal{O} \subset \mathcal{X}$ \ie there exists a finite $T>0$ such that \Review{$\mathcal{O} = \{\vec{x}(t) \: | \: \vec{x}(t) = \vec{\varphi}_T\left(\vec{x}\left( t \right)\right) = \vec{x}(t+T) \}$ with the exception for the horizontal position $q_x$.}{$\mathcal{O} = \{\vec{x}(t) \: | \: \vec{S}\vec{x}(t) = \vec{S}\vec{\varphi}_T\left(\vec{x}\left( t \right)\right) = \vec{S}\vec{x}(t+T) \}$, where $\vec{S} = diag(0,1,1,1...)$ is the selection matrix, which excludes the horizontal position $q_x$ from the periodicity.}{11-3}
\end{definition}
\vspace{3mm}

\Review{}{In this definition, we assume the evolution operator $\varphi_t$ is unique in both forward time ($t>0$) and backward time ($t<0$). }{11-2}
\Review{In order to generalize symmetries of gaits, we use group theory and define a symmetry as a group action that takes one gait and yields a gait of the same system:}{Given this assumption, we generalize the symmetries of gaits through group theory, defining a symmetry as a group action that transforms one gait into another within the same system:}{}
\begin{definition}[Symmetry]
\label{def:SYMGroup}
For any gait $\mathcal{O}$, the symmetries form a group $G$.
A symmetry of $G$ on the gait $\mathcal{O}$ is an assignment of a function $\mathcal{S}_g: \mathcal{O} \rightarrow \mathcal{O}$ to each element $g \in G$ in such a way that:
        \begin{itemize}
            \item If $I$ is the identity element of the group $G$, then $\mathcal{S}_I$ is the identity map, \ie for every gait $\mathcal{O}$ we have $\mathcal{S}_I(\mathcal{O}) = \mathcal{O}$.
            \item For any $g, h \in G$, we have $\mathcal{S}_g \circ \mathcal{S}_h = \mathcal{S}_{gh}$, \ie for every gait $\mathcal{O}$, we have $\mathcal{S}_g \left( \mathcal{S}_h(\mathcal{O})  \right) = \mathcal{S}_{gh}(\mathcal{O})$;
            \item For any $g \in G$, there exists the inverse of $g$ such that $\mathcal{S}_g \circ \mathcal{S}_{g^{-1}} = \mathcal{S}_{I}$;
        \end{itemize}
\end{definition}
\vspace{1mm}

\begin{figure}[tpb]
\vspace*{0.1in}
\centering
\includegraphics[width=1\columnwidth]{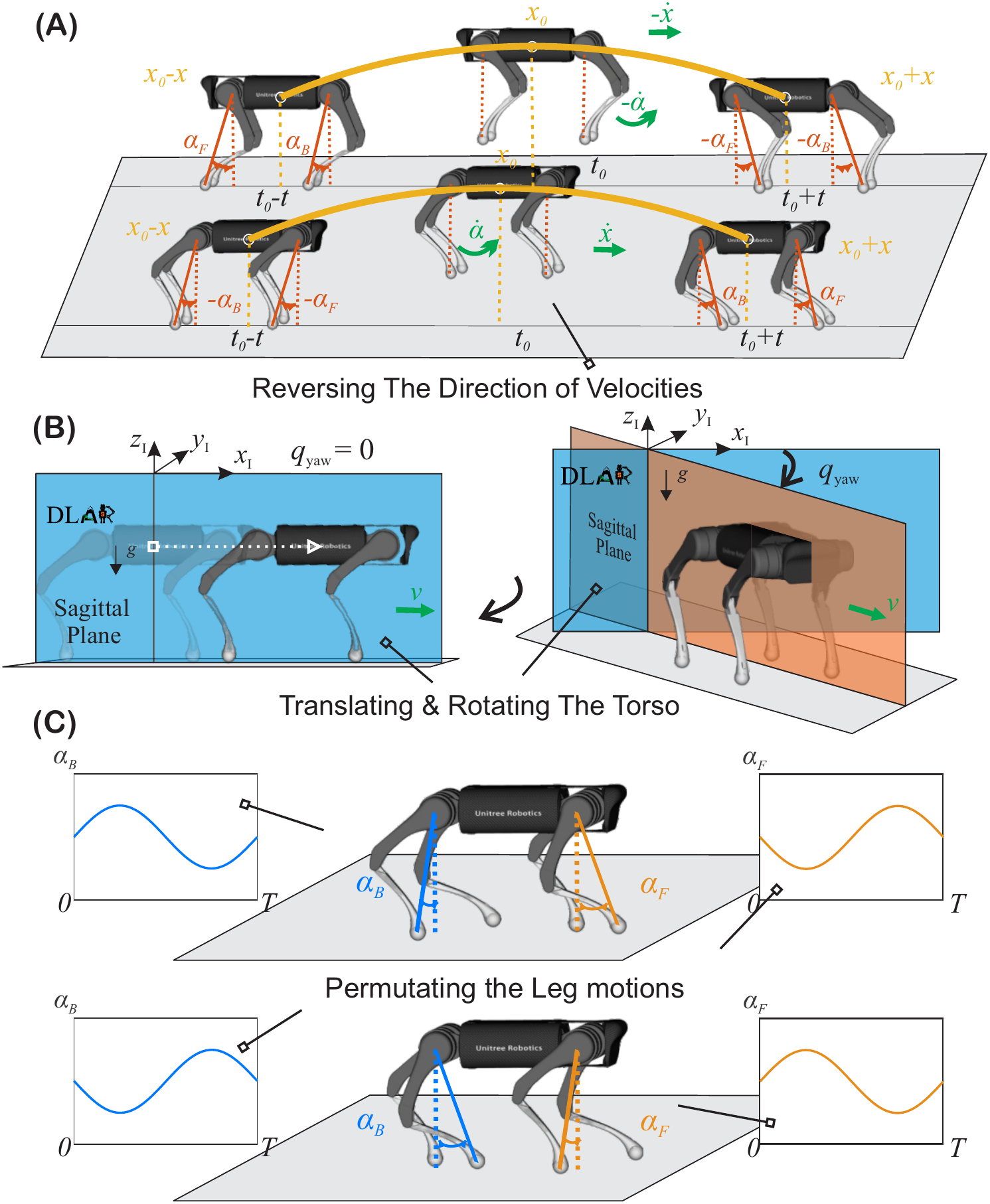}
\caption{
This figure illustrates three prevalent types of symmetries of a quadrupedal robot: (A) Time-Reversal Symmetry (a specific instance of Temporal Symmetry) denoted as $G_{\psi}$, wherein reversing all velocities yields a backward-moving gait. (B) Spatial Symmetry denoted as $G_\xi$, which corresponds to 2-dimensional Euclidean isometries within $SE(2)$ that involve translations and rotations of the robot on the moving plane. (C) Morphological Symmetry denoted as $G_{\sigma}$, representing permutations of leg motions that do not alter the periodic movements of the robots.
}
\vspace*{-0.2in}
\label{fig:symmetries}
\end{figure}

\vspace{3mm}
\subsubsection{Temporal Symmetry}
Since gaits are periodic, starting from any point on the orbit, all states will return to their original value after every $T$ seconds. 
This immediately gives us the first kind of \Review{}{discrete }{11-3}symmetry about time.
The periodicity of the states also suggests the conservation of linear and angular moment in each direction over every $T$ second.
\vspace{1mm}
\begin{definition}[Temporal Symmetry]
Suppose there are no energy losses related to the impact map $\Delta$, there exists a finite $T>0$ and specific timings $t = nT$ ($n \in \Z^+$), and the evolution operator $\vec{\varphi}_{t}$ forms a subgroup named temporal symmetry $\vec{\varphi}_{nT}$ such that $\vec{\varphi}_{t_1} \circ \vec{\varphi}_{t_2} = \vec{\varphi}_{t_1 + t_2}$ for all $t_1, t_2 \in nT$ and $\vec{\varphi}^{-1}_{t} = \vec{\varphi}_{-t}$, in which $-t$ corresponds to reversing the motion and rewinding time for every gait.
Specifically, assume a state vector on the orbit takes the form $\vec{x}(t) = \left[\vec{q}_{T}(t)^\intercal, \vec{q}_{L}(t)^\intercal, \vec{\dot{q}}(t)_{T}^\intercal, \vec{\dot{q}}(t)_{L}^\intercal  \right]^\intercal \in \mathcal{O}$, the temporal symmetry is a function $\mathcal{S}_{\varphi}: \vec{x}(t) \rightarrow \vec{x}(t)$:
\begin{align}
    \mathcal{S}_{\varphi}: & \left[\vec{q}_{T}(t)^\intercal, \vec{q}_{L}(t)^\intercal, \vec{\dot{q}}_{T}(t)^\intercal, \vec{\dot{q}}_{L}(t)^\intercal  \right] ^\intercal , \\ \notag
    \mapsto & \vec{\varphi}_{nT} \left[\vec{q}_{T}(t)^\intercal, \vec{q}_{L}(t)^\intercal, \vec{\dot{q}}_{T}(t)^\intercal, \vec{\dot{q}}_{L}(t)^\intercal  \right] ^\intercal
\end{align}
\end{definition}

\vspace{3mm}
Even though once a gait is identified, in theory, we can always rewind the time and find the inverse mapping of the evolution operator, in reality, this process is usually nonphysical and cannot be realized on the hardware.
Under the presence of damping, friction, or collisions, the reverse mapping requires the legged system to reverse the energy flow and absorb energy from its environment.
Additionally, when the projection $\Delta$ is non-invertible, the solutions of a hybrid system are often not unique in backward time.
However, it has been commonly observed that both humans and quadrupedal animals utilize similar gaits when moving both forward and backward \cite{Raibert1986symmetry}. Also, as a theoretical study, this is still a very interesting phenomenon for many idealized \Review{energy}{energetically}{11-5} conservative models, such as SLIP models proposed in the previous section.
Here we only consider a special instance when the energy in the system is reversible and the reset map is continuous \ie an identity map, then we have the following \Review{}{discrete }{11-3}symmetry:
\vspace{1mm}
\begin{definition}[Time-Reversal Symmetry]
\label{def:SYM_TimeReversal}
The time reversal symmetry is an action of $G_{\psi}$ on the gait $\mathcal{O}$ endowed with a function $\mathcal{S}_{\psi}$ that reverses the velocities:
\begin{align}
    \mathcal{S}_{\psi}: & \left[\vec{q}_{T}(t)^\intercal, \vec{q}_{L}(t)^\intercal, \vec{\dot{q}}_{T}(t)^\intercal, \vec{\dot{q}}_{L}(t)^\intercal  \right] ^\intercal , \\ \notag
    \mapsto & \vec{\varphi}_{nT} \left[\vec{q}_{T}(t)^\intercal, \vec{q}_{L}(t)^\intercal, -\vec{\dot{q}}_{T}(t)^\intercal, -\vec{\dot{q}}_{L}(t)^\intercal  \right] ^\intercal
\end{align}
\end{definition}

\vspace{3mm}
This symmetry implies that when a robot faces forward while moving backward, it is identical to moving forward in a time-reversed fashion.
This definition of symmetry has been discussed by Marc Raibert on his bipedal hopping robot and quadrupedal bounding robot \cite{Raibert1986symmetry}. 
The reverse motion also satisfies the same set of EoMs as the forward motion and one cannot decide whether the system is moving forward or backward just by looking at the motion pictures \cite{Lamb1998}.

\vspace{3mm}
\subsubsection{Spatial Symmetry}
The environment of the system also has huge effects on the symmetries of gaits.
As we modeled legged systems as FBMs in the Euclidean space, they are known for having symmetries to translations and rotations.
If there is no gravity in the environment and the robot is floating in space, every transformation $SE(3)$ of the torso's state is a symmetry that preserves the mass and inertia of the bodies and the linear/angular momentum of the robot.
As soon as the robot is subjected to gravity and constrained to move on a surface, it breaks the spatial symmetry of the system.
Here, for simplicity, we only consider the ground (moving surface) to be flat and perpendicular to the gravitational field.
\vspace{1mm}
\begin{definition}[Spatial Symmetry]
\label{def:SYM_SE2}
Given the assumptions above, the 
2 dimensional Euclidean isometries $SE(2)$ form a \Review{}{continuous }{11-3}symmetry subgroup $G_\xi \subset G$ of a gait $\mathcal{O}$.
An action of $G_\xi$ on the gait $\mathcal{O}$ is an assignment of a function $\mathcal{S}_{\xi}: \mathcal{O} \rightarrow \mathcal{O}$ to each element $\xi \in G_\xi$ in such a way that:
\begin{align}
    \mathcal{S}_{\xi}: & \left[\vec{q}_{T}(t)^\intercal, \vec{q}_{L}(t)^\intercal, \vec{\dot{q}}_{T}(t)^\intercal, \vec{\dot{q}}_{L}(t)^\intercal  \right] ^\intercal , \\ \notag
    \mapsto & \left[\mathcal{T}(\vec{q}_{T}(t))^\intercal, \vec{q}_{L}(t)^\intercal, \vec{\dot{q}}_{T}(t)^\intercal, \vec{\dot{q}}_{L}(t)^\intercal  \right] ^\intercal
\end{align}
where $\mathcal{T}$ is a linear transformation in $SE(2)$ that changes the torso's position $q_x, q_y$ and heading direction $q_{\text{yaw}}$ in the inertial frame. 
\end{definition}

This is the most obvious but often ignored symmetry in a legged system, which applies to any gaits.
It suggests a robot can start from anywhere in space with a certain height and move in any direction using the same gait without altering the leg movements.
This is also referred to as the group of linear transformations in mechanics and they represent changes in coordinates without affecting the formula for differential equations of motion \cite{singer2001symmetry}.
One interesting example based on this definition is that rotating the whole system in yaw $q_{\text{yaw}}$ will not change the motion (Fig.~\ref{fig:symmetries}(B)).
As for a planner system, it implies the model can also move in the opposite direction by changing $q_{\text{yaw}} = \pi$ rad in the inertial frame. 
This will be referred to as $\mathcal{S}_{\xi(\pi)}$ afterwords.
However, it is important to distinguish this symmetry from the temporal symmetry mentioned in the previous definition.
Even though in both cases, the robot is moving backward, the orientation $q_{\text{yaw}}$ of the robot is not changed in the temporal symmetry.

\vspace{3mm}
\subsubsection{Morphological Symmetry}
In a similar fashion to the morphologies of animals in nature, most of the robots also possess a bilateral symmetry (left and right sides are mirrored images of each other with respect to the sagittal plane) that reduces complexity in design, manufacturing, and control.
This symmetry will also introduce additional geometrical symmetries in gaits.
The detailed discussion of modeling of articulated leg configurations has been omitted due to space constraints.
In the following analysis, we assume that all legs have an identical structure and they are connected to the torso at the hip and shoulder joints such that the bilateral symmetry of the system with respect to the sagittal plane is retained (see the model in Fig1.B as an example).

\begin{definition}[Morphological Symmetry]
\label{def:LegSymmetry}
Assume a robot has the bilateral symmetry as described above, the permutations $\sigma$ of the index set of four legs $L = {\{\footnotesize\text{LH}, \text{LF}, \text{RF}, \text{RH}\}}$ form a \Review{}{discrete }{11-3}symmetry subgroup $G_{\sigma}$.
Morphological symmetry on the gait $\mathcal{O}$ is a function $\mathcal{S}_{\sigma}$:
\begin{align}
    \mathcal{S}_{\sigma}: & \left[\vec{q}_{T}(t)^\intercal, \vec{q}_{L}(t)^\intercal, \vec{\dot{q}}(t)_{T}^\intercal, \vec{\dot{q}}(t)_{L}^\intercal  \right] ^\intercal , \\ \notag
    \mapsto & \left[\vec{q}_{T}(t)^\intercal, \sigma(\vec{q}_{L}(t))^\intercal, \vec{\dot{q}}_{T}(t)^\intercal, \sigma(\vec{\dot{q}}_{L}(t))^\intercal  \right] ^\intercal
\end{align}
\end{definition}

\vspace{2mm}
Symmetries of leg permutations are abundant in quadrupedal gaits and they were extensively discussed in \cite{Hildebrand1965}.
During a stride, these symmetries suggest that the legs move in the same manner and contribute equally to the dynamics of the system (Fig.~\ref{fig:symmetries}(C)). 
There are various ways in which these symmetries can be broken, for example, when legs are no longer identical or when the location of the COM is closer to one leg pair.
It is demonstrated in the following section that by altering one of the system parameters, e.g., the total energy, the discrete morphological symmetry can be disrupted, potentially leading to unsynchronized leg movements and a novel gait characterized by diverse footfall patterns.

\section{Results}
\label{sec:results}
This section showcases our primary discoveries and illustrates how different parameter bifurcations break symmetries in a simplistic model in distinct ways, leading to various quadrupedal gaits and footfall patterns.
In particular, we employed numerical continuation techniques outlined in our earlier research \cite{Gan2018allcommonbipedal} and sought out periodic solutions for the model defined in equation \eqref{eq:HM}.
In this study, we revealed four unique quadrupedal gaits (pronking, bounding, half-bounding, and galloping) that demonstrate interconnections through diverse parameter bifurcations.
The naming conventions of several common quadrupedal gaits were adopted from the animal locomotion literature using the footfall patterns \cite{Hildebrand1965, Alexander1984bipelquadrupedal}.
In order to generalize our results for systems with different sizes, we normalized all the values in our results using total mass $m$ of the entire system, resting leg length $l_o$, and gravity $g$. 
\Review{In order to have an energy-conservative system and avoid collision losses, we took the limit of the foot mass $m_{o}$ to zero so that $m = M + 4 m_o = M$.}{In addition, we took the limit of the foot mass $m_{o}$ to zero so that $m = M + 4 m_o = M$ to have an energy-conservative system and avoid collision losses.}{}
For simplicity, most of the system parameters were fixed in this study with $k_{\text{leg}} = 10~[\nicefrac{mg}{l_o}]$, $\omega_{\text{swing}} = 20~[\sqrt{\nicefrac{g}{l_o}}]$, and $I = 2~[ml_o^2]$ according to the values used by animals as described in \cite{Blickhan1993, Polet2021, DingJerboa2022}.

To better understand the symmetry-breaking process, we started the search with a quadrupedal model characterized by the largest set of morphological symmetries \ie all four legs were identical and the COM was at the geometric center of the torso.
Then we broke various types of symmetries described in the method section by varying the total energy $E(\vec{q}, \dot{\vec{q}})$ stored in the system and the COM location $l_{b}$ of the torso and analyzed their influences on the footfall patterns of periodic gaits.
To simplify the numerical calculations and the analysis of the periodic orbits, we selected the apex transition in the flight phase as the Poincare section $\mathcal{P} = \{(\vec{q},\vec{\dot{q}}) \in \mathcal{TQ} \:|  ~\dot{q}_z = 0, ~\ddot{q}_z < 0, \: (t^{\text{TD}}_{i}(\vec{q},\vec{\dot{q}}) \mod{T}) <(t^{\text{LO}}_{i}(\vec{q},\vec{\dot{q}}) \mod{T})  ~\}$.
In the following figures, the fixed points $\mathcal{P}^* \de \mathcal{P}(t) = \mathcal{P}(t+T)$ on the Poincare section \ie the periodic orbits $\mathcal{O}$ are visualized using the torso's horizontal velocity $\dot{q}_x$ and pitching velocity $\dot{q}_{\text{pitch}}$.
This system has been developed in MATLAB, and the source code is available for download from our GitHub repository \footnote{\href{https://github.com/DLARlab/BreakingSymmetryLeadstoDiverseGaits}{https://github.com/DLARlab/BreakingSymmetryLeadstoDiverseGaits}}.
Please also refer to the multimedia file or visit the web page \footnote{\href{https://dlarlab.syr.edu/research/breaking-symmetries/}{https://dlarlab.syr.edu/research/breaking-symmetries/}} to view animations of the gaits in this study.

\subsection{Pronking and Bounding Gaits}
\label{subsec:Energy}

\subsubsection{Pronking \textup{(PF)}} 
\begin{figure}[tp!]
\centering
\includegraphics[width=1\columnwidth]{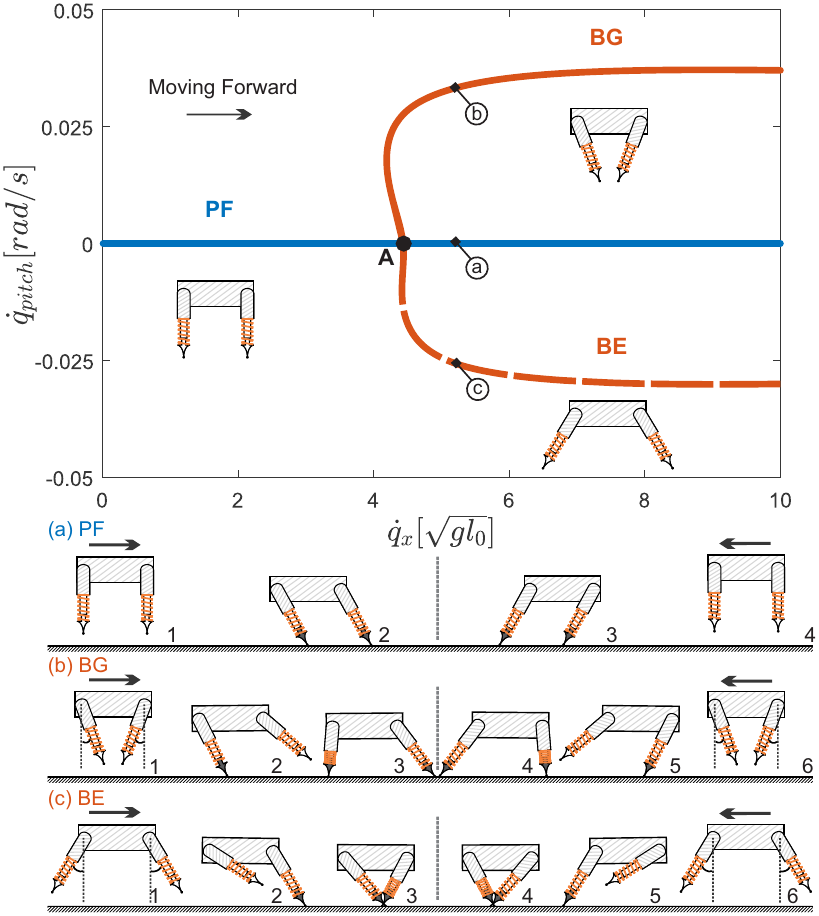}
\caption{Gait branches of pronking forward (PF), bounding with gathered suspension (BG), and bounding with extended suspension (BE) on the Poincare section with the torso's horizontal velocity $\dot{q}_x$ as the horizontal axis and pitching velocity $\dot{q}_{\text{pitch}}$ as the vertical axis. Examples of each gait (a-c) are shown as successive keyframes at the touch-down, lift-off moments, and the apex at the bottom. Black feet are used to highlight the legs in stance.
}
\label{fig:PronkBound}
\vspace*{-0.2in}
\end{figure}

is a quadrupedal gait frequently observed in quadrupedal animals, characterized by the synchronized movement of all four legs.
During a single stride, there is only one flight phase followed by a stance phase, during which all four legs make contact with the ground, moving in precisely the same manner.
This results in zero torque on the torso and the torso has no rotational motion throughout the stride ($\dot{q}_{\text{pitch}} = 0~[rad/s]$).
Our parameter continuation process initiated with a simple seed solution, where the model executed an in-place jump, with all four legs aligned vertically downward ($\dot{q}_x = 0~[\sqrt{gl_o}]$ and $\dot{q}_{\text{pitch}} = 0~[rad/s]$).
When the overall energy in the system was modified, the resulting solutions constituted a one-dimensional curve (blue curve in Fig.~\ref{fig:PronkBound}) with varying average speeds.
Multiple consecutive keyframes of the pronking gait, starting with an initial speed of $5.2~[\sqrt{gl_o}]$, at the instances of the apex, touchdown, and liftoff, were illustrated in Figure~\ref{fig:PronkBound}(a).
In our definitions, the pronking gait exhibits the largest number of symmetries. It is characterized by the Time-Reversal Symmetry $\mathcal{S}_{\psi}$, which means that by reversing the signs of the velocity states, one can immediately identify another periodic motion without changing the configurations or requiring additional numerical integration. As shown in Figure~\ref{fig:PronkBound}(a), this implies that examining the keyframes of the pronking gait in the sequence of 4-3-2-1 also represents a pronking gait moving in reverse with a negative horizontal speed. Furthermore, it is worth noting that due to the uniformity of the leg motions, the pronking gait preserves all morphological symmetry related to leg permutations $\mathcal{S}_{\sigma}$, where $\sigma \in \mathbb{S}_L$ ( $\mathbb{S}_L$ denotes all permutations of the set $L$).

\subsubsection{Bounding \textup{(BG and BE)}} Compared to pronking gaits, bounding gaits exhibit a breakdown in the coordination between the front and hind legs.
Despite the synchronized movements within the front and hind leg pairs, there is a phase delay between legs on the same sides, leading to two touchdown and liftoff events within a single stride.
Our model yielded two distinct types of solutions that adhere to this pattern.
In the first type of bounding gait, as depicted in Fig.~\ref{fig:PronkBound}(b), the hind leg pair makes initial contact with the ground, followed by the front legs, causing the leg pairs to converge inward during the flight phase. This particular gait is referred to as \emph{bounding with gathered suspension}, as described in \cite{Hildebrand1977}, and we'll abbreviate it as BG hereafter.
In Fig.~\ref{fig:PronkBound}, these solutions are represented by the solid orange curve, which bifurcates from the PF branch at $\dot{q}_x = 4.4~[\sqrt{gl_o}]$ (designated as black dot A).
The other type of bounding gait is characterized by the opposite sequence of touch-downs, as shown in Fig.~\ref{fig:PronkBound}(c), where the front legs touch the ground first, followed by the hind legs. 
Consequently, the swing leg pairs extend outward during the flight phase, termed \emph{bounding with extended suspension} (BE).
In Fig.~\ref{fig:PronkBound}, these solutions are represented by red dashed curves, connected to the pronking branch at the same bifurcation point A.

Although the footfall patterns and swing leg behaviors of these two bounding gaits differ, they share common underlying symmetries.
Both BG and BE possess an equal number of morphological symmetries $\mathcal{S}_{\sigma}$, where $\sigma \in \{\text{(LF,RF),~(LH,RH),~(LF,RF)(LH,RH)}\}$.
This means that the system states of legs within the front or hind leg pairs follow identical time series. 
Consequently, altering the states within one or both leg pairs will not impact the system's dynamics.
Contrasting with the PF gait discussed in the previous section, the distinguishing feature in the bounding gaits lies in the symmetry breakdown between the front and back leg motions, represented by $\sigma =$ (LH,LF)(RF,RH). 
This introduces a phase lag in the leg pairs, which can be either positive or negative, giving rise to two new gaits: BG and BE, originating from point A.
Furthermore, it is noteworthy that new periodic orbits can be identified for both BG and BE gaits by merely reversing all the speeds while retaining the same system configurations.
This highlights the presence of time-reversal symmetry $\psi$.
Without performing an additional numerical search, the keyframes displayed in Fig.~\ref{fig:PronkBound} (b)\&(c) represent two distinct periodic solutions with opposite moving directions: one advances from the 1st frame to the 6th frame with a positive speed, while the other moves in the reverse direction, from the 6th frame back to the 1st frame with a negative speed.

\subsection{Pronking and Bounding Gaits of an Asymmetric Model}
\label{subsec:COM}
The time-reversal symmetry $\psi$ observed in the pronking and bounding gaits illustrated in Fig.~\ref{fig:PronkBound} offers a practical method for identifying two periodic motions within a single numerical search.
Nevertheless, as demonstrated in this subsection, the presence of this symmetry is strongly contingent on the morphological symmetry concerning the frontal plane, where the mechanical components of the system's anterior and posterior are mirrored.
To be more precise, our findings indicate that when the COM is shifted away from the body's central point, the pronking and bounding gait branches discussed in the previous section undergo rapid changes.

\begin{figure}[tp!]
\centering
\includegraphics[width=0.96\columnwidth]{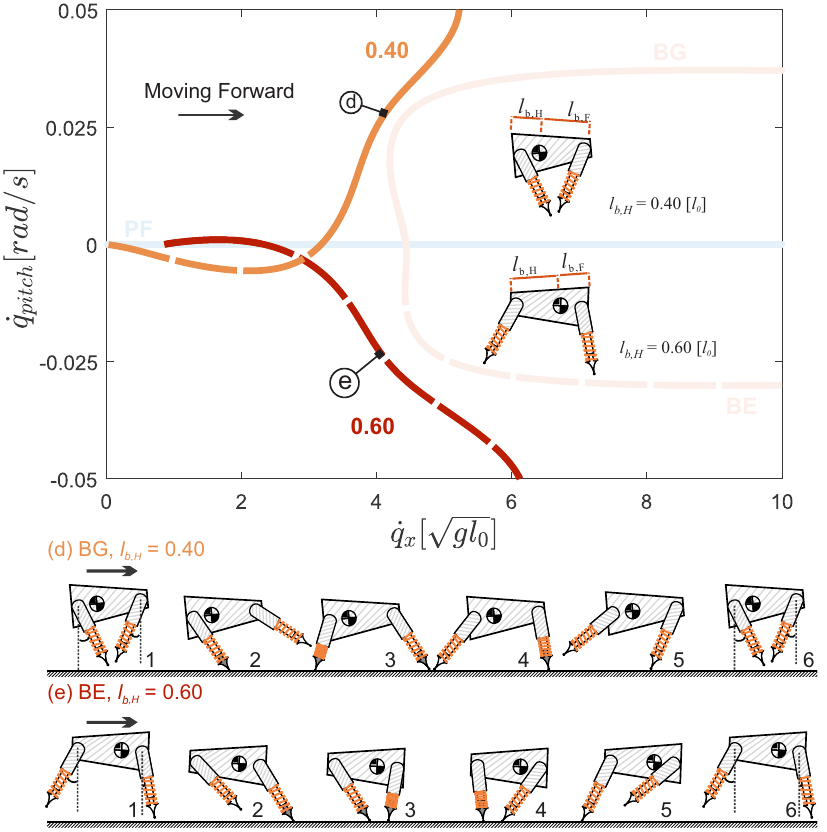}
\caption{The figure displays bounding gait branches for asymmetrical models where $l_{b,H} \neq 0$. Two specific cases, $l_{b,H}=0.40~[l_o]$ and $l_{b,H}=0.60~[l_o]$, are highlighted, with their corresponding solution branches illustrated in orange and red, respectively. Solid lines represent bounding gaits with gathered suspensions (BG), while dashed lines depict bounding gaits with extended suspensions (BE). In contrast to the bounding gaits of the symmetrical model shown in Fig.~\ref{fig:PronkBound}, our numerical analysis indicates that as the COM shifts closer to the rear, BG manifests at intermediate speeds (d). In contrast, when the COM is nearer to the front at $l_{b,H} = 0.60~[l_o]$, BE remains the exclusive bounding gait.
}
\label{fig:PronkBound_Asym}
\vspace*{-0.23in}
\end{figure}

We conducted numerical continuations using two models with distinct parameters, where $l_{b,H}$ was set at $0.4~[l_o]$ for the model with the COM closer to the posterior, and $0.6~[l_o]$ for the model with the COM closer to the anterior.
To clearly differentiate the results, we utilize a trapezoidal shape to represent the robot's torso, positioning the COM closer to the side with the longer base.
During forward motion ($\dot{q}_x > 0$), the pronking branch PF is no longer present in both models.
This absence is attributed to the differing lever arm lengths of the four legs, preventing the leg forces from producing zero torque on the torso during the stance phases when all four legs were fully synchronized.
In each of the models, we observe only one type of bounding gait at a given speed.
In the model with $l_{b,H} = 0.4~[l_o]$ (as shown in Fig.~\ref{fig:PronkBound_Asym}(d)), BE appears within lower speed ranges (indicated by the dashed orange curve), while BG emerges (solid orange curve) as the exclusive gait at speeds exceeding $\dot{q}_x = 3.2~[\sqrt{gl_o}]$.
Conversely, in the model with the reversed COM shift (depicted in Figure~\ref{fig:PronkBound_Asym}(e)), BE (represented by the dashed red line) is the sole bounding gait seen at speeds surpassing $\dot{q}_x = 2.5~[\sqrt{gl_o}]$.
In comparison to the symmetrical model discussed in the previous section, both bounding gait branches are only present within intermediate speed ranges and vanish rapidly when the speed exceeds $\dot{q}_x = 5~[\sqrt{gl_o}]$.

However, it is important to observe that, as depicted in Fig.~\ref{fig:PronkBound_Asym}(d)\&(e), the pitch angles are no longer zero, and the magnitudes of leg angles differ during the apex transition in the first and last frames.
This indicates that the time-reversal symmetry $\psi$ is no longer preserved for a gait originating from a model with $l_{b,H} \neq 0.50~[l_o]$.
Hence, reversing the speeds (via $\psi$) does not result in discovering gaits that moved backward \ie observing the frames in the sequence of 6-1 does not represent a periodic motion for the same model.
Furthermore, for a given speed, the model with full symmetry illustrated in Fig.~\ref{fig:PronkBound} has two feasible bounding gaits readily available.
In contrast, for the asymmetrical model depicted in Fig.~\ref{fig:PronkBound_Asym}, only one bounding gait exists, and the viable footfall pattern depends on the location of the COM.

\subsection{Half-Bounding and Galloping Gaits}
\subsubsection{Half-Bounding \textup{(FG, FE, HG, and HE)}} 
\begin{figure}[tbp]
\centering
\includegraphics[width=1\columnwidth]{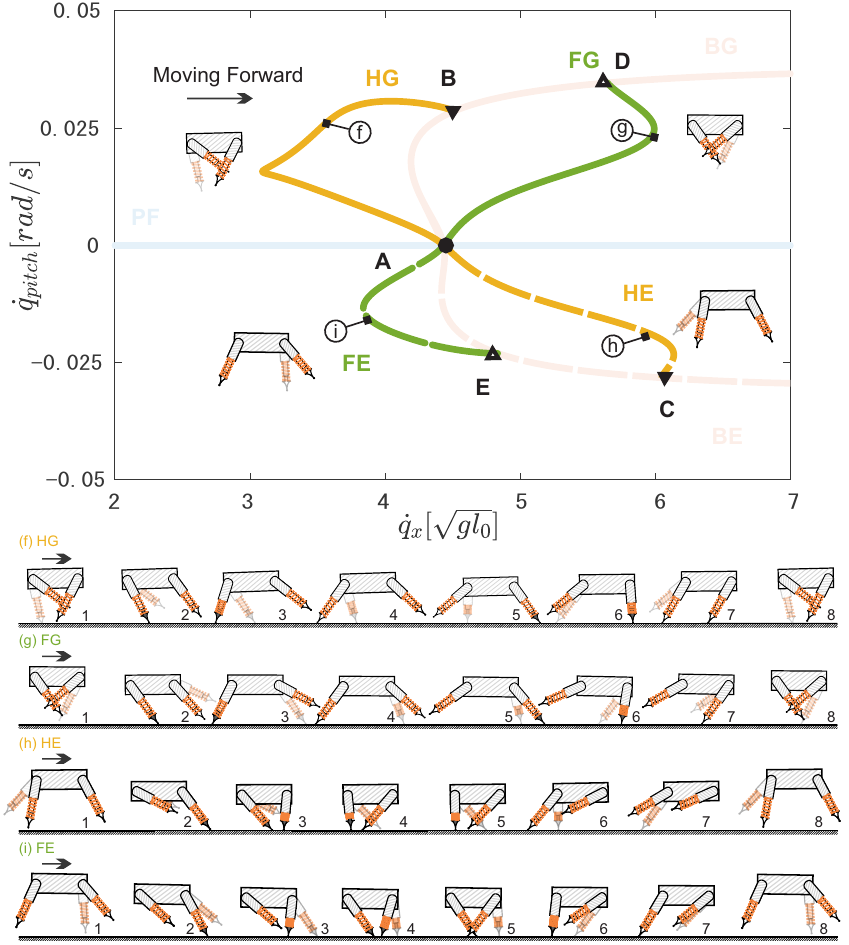}
\caption{This figure illustrates the gait branches of half-bounding as identified from the proposed model. Since symmetry breaking can occur in either the front or hind leg pair, during the numerical search, a total of four such gait patterns were discovered from the two bounding gaits (BG and BE):
(f) and (g) demonstrate half-bounding gaits with gathered suspensions;
(h) and (i) illustrate half-bounding gaits with extended suspensions.
}
\label{fig:HalfBounding}
\vspace*{-0.2in}
\end{figure}

are similar to bounding gaits but the synchronization is broken in one leg pair.
In these gaits, one of the leg pairs makes contact with the ground simultaneously, while the other leg pair touches down in rapid succession.
Consequently, half-bounding gaits can be classified as 3-beat gaits.
We found in total four variations of them based on the footfall patterns as shown in Fig.~\ref{fig:HalfBounding}.
When this desynchronization occurred in the hind leg pair, we found “half-bounding with spread \textbf{H}ind legs and \textbf{G}athered suspension" (HG) and “half-bounding with spread \textbf{H}ind legs and \textbf{E}xtended suspension" (HE).
Similar to the distinction between bounding gaits in BG and BE branches, solutions within the FG and FE branches exhibit varying leg orientations and pitching velocities during the apex transitions.
For instance, as depicted in Fig.~\ref{fig:HalfBounding}, the HG gait branch (yellow solid curve) emerges from point B on BG at a speed of $\dot{q}_x = 4.6~[\sqrt{gl_o}]$, while the HE branch (yellow dashed curve) originates at point C on BE with a speed of $\dot{q}_x = 6.1~[\sqrt{gl_o}]$.
As the solutions along the HG and HE branches approach point A, all four legs tend to synchronize, ultimately converging with the pronking branch at point A.
The keyframes of periodic solutions for these two gaits at the moments of touchdown, liftoff, and the apex transition are illustrated in Fig.~\ref{fig:HalfBounding}(f)\&(h).
On the other hand, when the symmetry breaking occurred in the front leg pair, we identified two additional half-bounding gaits characterized by front legs in a spread position and gathered suspension (FG) and front legs in a spread position and extended suspension (FE). 
These two branches are represented as green solid/dashed curves in Fig.~\ref{fig:HalfBounding}, and they connect to the bounding gaits BG and BE at bifurcation points D and E, respectively, with forward speeds of $5.7~[\sqrt{gl_o}]$ and $4.8~[\sqrt{gl_o}]$, correspondingly. 
Similarly, the FG and FE branches ultimately converge with the pronking branch and merge at bifurcation point A. 
Keyframes illustrating these two gait solutions can be observed in Fig.~\ref{fig:HalfBounding}(g)\&(i).
The break in symmetry in one leg pair causes phase delays that ultimately lead to the half-bounding gaits with different leading and trailing legs, which resemble the bipedal skipping gaits discussed extensively in our previous work \cite{GanDynamicSimilarity}.
During this process, only one morphological symmetry is retained in the half-bounding gaits \ie $\sigma =$ {\footnotesize (LF,RF)} or {\footnotesize (LH,RH)} and the time-reversal symmetry $\psi$ ceases to exist.

\subsubsection{Galloping} 
\begin{figure}[tbp]
\centering
\includegraphics[width=1\columnwidth]{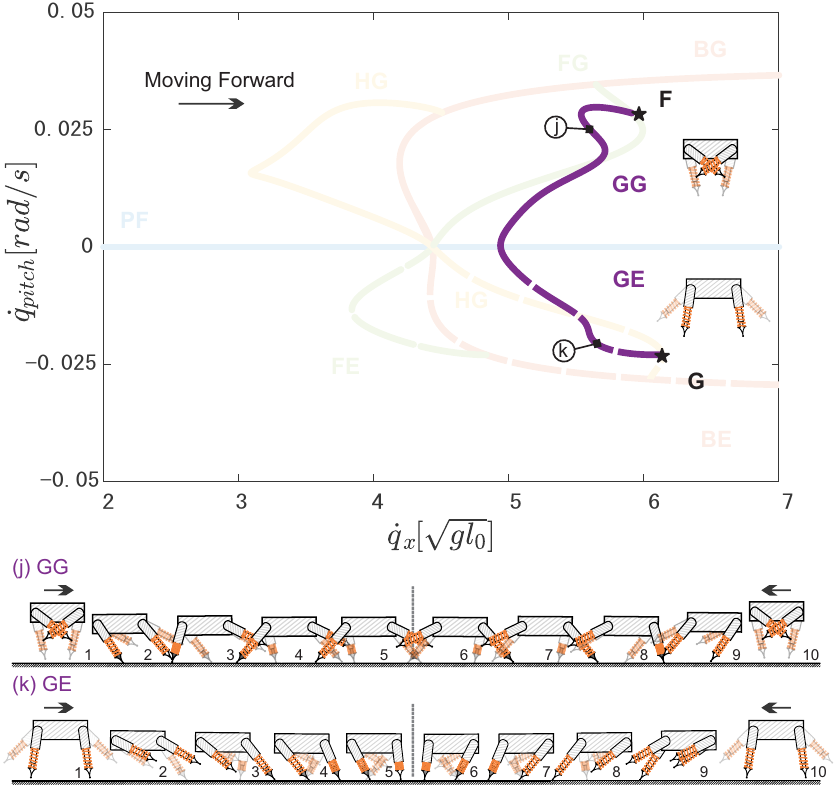}
\caption{This figure displays the solution branch and keyframes for two galloping gaits:
(j) Galloping with gathered suspension (GG);
(k) Galloping with extended suspension (GE).
In this visualization, the left legs are presented in a transparent color, while black feet are employed to emphasize the legs in the stance phase.}
\label{fig:Galloping}
\vspace*{-0.2in}
\end{figure}

In the study conducted by Hilderbrand (1989) \cite{Hilderbrand1989Quadrupedal}, a range of galloping gaits was identified, each characterized by unique footfall patterns and aerial suspension.
In our research, we analyzed galloping patterns from the proposed model and discovered two distinct types: “galloping with gathered suspension" (GG) and “galloping with extended suspension" (GE).
These patterns are represented by purple curves in Fig.\ref{fig:Galloping}, connecting to the FG and HE branches at bifurcation points F and G, marked as stars, with speeds of $\dot{x} = 6.0~[\sqrt{gl_o}]$ and $\dot{x} = 6.2~[\sqrt{gl_o}]$ respectively.
Similar to the bounding gaits, these galloping gaits exhibit unique touchdown sequences and swinging leg motions, reminiscent of the galloping movements observed in natural horses and gazelles.
It is worth noting that, as per our definition, galloping represents the least symmetrical gait due to the absence of all symmetries in leg permutations.
Despite this lack of symmetry, solutions that retain time-reversal symmetry $\psi$ are still present, as depicted in Fig.~\ref{fig:Galloping}(j)\&(k).
Due to space constraints, this study does not distinguish between the right or left leading legs in galloping gaits. Observations from animal locomotion indicate that altering the motion of the front legs can result in either a ``rotary gallop" or a ``transverse gallop" \cite{bertram2009motions} which is another symmetry that is not discussed here.

\section{Conclusions}

In this study, we conducted a comprehensive exploration of the symmetries in quadrupedal legged locomotion, employing the principles of group theory.
First, we established a systematic framework for defining gaits as periodic solutions of a hybrid system, allowing us to categorize all the observed symmetries in legged animals and robots into three primary subgroups: temporal symmetry denoted as $\psi$; spatial symmetry represented by $\xi$; and morphological symmetry $\sigma$.
In this framework, we found that any composition of these symmetries still constitutes a symmetry in the context of quadrupedal locomotion.
Consequently, the various quadrupedal gaits can be effectively distinguished and predicted by quantifying the number of elements within the symmetry group, thus shedding light on the rich and diverse repertoire of various footfall sequences of quadrupedal legged locomotion.

Secondly, to gain a comprehensive understanding of the intricate relationships inherent in quadrupedal gaits, we developed a simplistic model and conducted numerical bifurcations.
This approach allowed us to visually depict how the disruption of symmetries can precipitate changes in the existence of gaits within a legged system.
More specifically, our research revealed that the pronking gait exhibits the highest number of symmetries, with other gaits such as bounding, half-bounding, and galloping branching out in a tree-like structure from one to another:
Stemming from the pronking gait, the desynchronization of motion between two leg pairs results in the emergence of two bounding gaits with gathered or extended suspensions, as illustrated in Fig.~\ref{fig:PronkBound}.
When only one leg pair initiates movement out of phase, it simultaneously disrupts both leg permutation and time-reversal symmetries, leading to the discovery of four distinct half-bounding gaits, as depicted in Fig.~\ref{fig:HalfBounding}.
These solutions were identified using a single energy-conserving model without the need for additional control laws or actuation. To some degree, they represent distinct oscillation modes within the same hybrid system, triggered solely by initial conditions like forward speed and the torso's height. This work not only provides crucial insights into the rationale behind utilizing multiple gaits at varying speeds but also holds the promise of an efficient and versatile strategy for generating reference trajectories for robotic systems with desired footfall sequences.

\section{Acknowledgement}
The authors are grateful for the valuable suggestions provided by Dr. Amit K. Sanyal from the Mechanical and Aerospace Engineering Department at Syracuse University.

\bibliographystyle{IEEEtran}
\bibliography{Reference}

\end{document}